\documentclass[11pt,a4paper]{article}

\usepackage[utf8]{inputenc}
\usepackage{amsmath}
\usepackage{amssymb}
\usepackage{amsthm}
\usepackage{mathtools}
\usepackage{geometry}
\usepackage{graphicx}
\usepackage{float} 
\usepackage{booktabs} 
\usepackage{algorithm}
\usepackage{algpseudocode}
\usepackage{hyperref}

\usepackage{bm}

\hypersetup{
    colorlinks=true,
    linkcolor=blue,
    filecolor=magenta,      
    urlcolor=cyan,
    pdftitle={Semantic Faithfulness and Entropy Production Measures for LLM Evaluation},
    pdfauthor={Igor Halperin},
}

\newcommand{\matA}{\mathbf{A}}
\newcommand{\matQ}{\mathbf{Q}}

\newcommand{\beq}{\begin{equation}}
\newcommand{\eeq}{\end{equation}}
\newcommand{\bea}{\begin{eqnarray}}
\newcommand{\eea}{\end{eqnarray}}

\def\argmax{\text{argmax}}
\def\arg{\text{arg}}

\def\boldmu{\text{\boldmath$\mu$}}

\def\boldnu{\text{\boldmath$\nu$}}

\def\boldlambda{\text{\boldmath$\lambda$}}
\def\boldxi{\text{\boldmath$\xi$}}

\def\Sdot{\overset{\bm .}{S}}

\title{Semantic Faithfulness and Entropy Production Measures 
\\
to Tame Your LLM Demons and Manage Hallucinations}

\author{Igor Halperin\thanks{The author acknowledges the assistance of Claude Code and Claude Sonnet 4.5 in developing code, generating and analyzing data, and preparing this manuscript.
All remaining errors are the author's own. The views expressed herein are those of the author and do not necessarily reflect the views of his employer. Code and data available at: \url{https://github.com/ighalp/semantic-faithfulness-sdm}. Email for correspondence: ighalp@gmail.com.} \\ Fidelity Investments}

\date{\today}

\begin{document}

\maketitle

\begin{abstract}
Evaluating faithfulness of Large Language Models (LLMs) to a given task is a complex challenge. We propose two new unsupervised metrics for faithfulness evaluation using insights from information theory and thermodynamics. Our approach treats an LLM as a bipartite information engine where hidden layers act as a Maxwell demon controlling transformations of context $ C $ into answer $ A $ via prompt $ Q $.
We model Question-Context-Answer (QCA) triplets as probability distributions over shared topics. Topic transformations from $ C $ to $ Q $ and $ A $ are modeled as transition matrices $\matQ$ and $\matA$ encoding the query goal and actual result, respectively. Our semantic faithfulness (SF) metric quantifies faithfulness for any given QCA triplet by the Kullback-Leibler (KL) divergence between these matrices. Both matrices are inferred simultaneously via convex optimization of this KL divergence, and the final SF metric is obtained by mapping the minimal divergence onto the unit interval [0,1], where higher scores indicate greater faithfulness.
Furthermore, we propose a thermodynamics-based semantic entropy production (SEP) metric in answer generation, and show that high faithfulness generally implies low entropy production. The SF and SEP metrics can be used jointly or separately for LLM evaluation and hallucination control. We demonstrate our framework on LLM summarization of corporate SEC 10-K filings.
\end{abstract}

\section{Introduction}

Large Language Models (LLMs) have demonstrated remarkable capabilities in generating human-like text, answering questions, and summarizing information. However, ensuring that their outputs are factually grounded and faithful to a provided source context is a persistent and crucial challenge. Large deviations of LLM outputs
from those expected according to their prompts, often referred to  
as {\it faithfulness hallucinations}, pose a major barrier to the deployment of LLMs in high-stakes domains such as medicine, law, and finance.

Current evaluation methods often rely on human annotators, which is expensive and not scalable, or on other LLMs for assessment, which can introduce its own biases and inaccuracies. There is a pressing need for automated, quantitative, and interpretable metrics that can reliably measure the faithfulness  of an LLM's response to a given context.

In this paper, we introduce a novel approach to this problem using insights from information theory \cite{csiszar2004, Cover_IT_book} and thermodynamics \cite{Parrondo_2015, Seifert_2012}. 
First, we model a triplet $(Q,C,A) $ of context $ C $, question $ Q $, and answer $A $ not as simple strings of text, but as distributions over a latent topic space. The process of transforming the initial text $ C$ into the final LLM output $ A $ can then be thought of as a transition between the initial topic distribution $ {\bf p}_{c} $ of the context document into the final topic distribution $ {\bf p}_{a} $. We can write such relation as 
$ {\bf p}_{a} = {\bf p}_{c}^{T} {\bf A} $, 
where $ {\bf A} $ is a $ N \times N $ matrix (where $ N $ is the number of topics) with all elements being non-negative and $ \sum_{j=1}^{N} {\bf A}_{ij} = 1, \; \forall i $. 
We can now similarly introduce a 'goal interpretation' matrix $ {\bf Q} $ that should be consistent with the marginal distributions $ {\bf p}_{c} $ and $ {\bf p}_{q} $, via the constraint 
$ {\bf p}_{q} = {\bf p}_{c}^{T} {\bf Q} $. The matrix 
$ {\bf Q} $ should also satisfy the probability constraint
$ \sum_{j=1}^{N} {\bf Q}_{ij} = 1, \; \forall i $.

 The core idea of our approach is that a "faithful" answer should transform the topics from the context in a way that is semantically similar to how the question "queries" the topics from that same context. We formalize this intuition by defining two probabilistic transition matrices $ {\bf A} $ and $ {\bf Q} $ and propose minimizing their Kullback-Leibler (KL) divergence as an objective. The resulting minimal divergence value serves as our faithfulness  score which we will call semantic faithfulness (SF) score. The SF metric is unsupervised, computable for a single (Question, Context, Answer) triplet, and provides a continuous measure of faithfulness, allowing for direct comparison between different LLM responses.
 
While the SF metric is based on the information-theoretic analysis, in this paper we also pursue a view of LLMs as {\it information engines}, and address thermodynamics of such engines \cite{Parrondo_2015, Seifert_2012}.  
We conceptualize a LLM as a {\it bipartite} information engine made of two sub-systems $ X $ and $ Y $. Only one of these sub-systems ($ X $) is partially observed to the user via LLM's API, while the second, unobserved sub-system $ Y $ conceptualizes the computational engine of the LLM. In the physics literature, such unobserved sub-systems of information engines are often referred to as {\bf Maxwell's demons}, and we will follow this terminology in this paper. 

The thermodynamics-based analysis of the LLM as an information engine produces our second metric for evaluation of faithfulness and managing LLM hallucinations, which we call semantic entropy production (SEP). To the extend that LLM hallucinations can be thought of as noisy (i.e. entropy-increasing) distortions of information contained in context $ C$ and user question $ Q $, the concept of entropy production as defined in physics \cite{Parrondo_2015, Seifert_2012} offers a quantitative way to quantify such measure in the LLM answer generating process.
We believe that searching for LLM hallucinations using the semantic entropy production metric is a more theoretically grounded and empirically attractive idea than looking only into the marginal semantic entropy of LLM answers, as is done in the Semantic Entropy method \cite{farquhar2024semantic}, see below in Sect.~\ref{sect_SE_or_SEP}. 
  
As we will show below, the SEP metric is easily computable using a part of our first algorithm for computing the SF metric. In certain cases discussed below, the SEP metric does not require any new optimization, and the SEP score can be computed directly off the SF score. Moreover, we will show that the SF and SEP scores are typically inversely related: a high SF score (high faithfulness) typically implies a low SEP score, and vice versa. This observation gives support to our SF metric, as well to the intuitive idea that faithfulness hallucinations can be thought of as LLM responses with extremely low semantic faithfulness.

We validate our framework experimentally on 10 Question-Context-Answer triplets from NVIDIA's fiscal 2024 10-K disclosure, organized into two groups testing different question structures: comprehensive multi-topic risk analysis versus focused competitive threats analysis. Our results demonstrate that the proposed metrics successfully capture meaningful differences in semantic faithfulness, with question structure, and not just entropy magnitude, emerging as the key driver of faithfulness variation. LLM-as-a-Judge evaluation~\cite{zheng2023judging} confirms that higher $\mathcal{F}_S$ scores correlate with superior structural alignment and contextual grounding.

The paper is organized as follows.
In Sect.~\ref{sec:metric} we present our problem formulation and modeling framework, and define the semantic faithfulness (SF) metric.
In Sect.~\ref{sect_AM_algorithm}, we present a lightweight numerical algorithm to compute the SF metric using the standard convex optimization software. In Sect.~\ref{sect_Entropy_Production}, we model the LLM as a thermodynamics engine and compute the SEP metric.
Sect.~\ref{sec:experiments} presents our experiments, and the final Sect.~\ref{sect_conclusions} concludes.           

\section{Semantic Faithfulness}
\label{sec:metric}

\subsection{Problem formulation}
\label{sect_problem}

In this paper, we consider a user interaction with a LLM viewed as a black box, and operated at the level of a single-step interaction when the user asks a question $ Q $ about some context document set $ C $ (e.g. to provide a summary of $ C $), and gets an answer $ A $ from the LLM. 
The black-box setting of our formulation implies that we do not have access to additional information such as internal activations, log-probs etc.  

In practice, tasks such as a LLM-provided summarization are often ran a few (say, $ K = 10 $) times, and then the 'best' answer is selected from $ K $ candidate answers by analysing the resulting triplets $ (Q_k, C, A_k) $, with $ k = 1, \ldots, K $. 
Instead of running the same question (prompt) $ K $ times, we assume a more general setting where all questions $ Q_k $ are constructed as semantically equivalent paraphrases of a given initial question $ Q $. As discussed in \cite{halperin2025sdm}, prompt paraphrasing is useful to enrich the data and extract 
more meaningful topic representations.
In what follows, we assume that triplets $ (Q_k, C, A_k) $ constitute the only available data for our analysis. 

The main question here is of course how to define what we mean by the 'best' answer. While multiple empirical schemes can be considered at this point, in general, 'the best' means 'the most faithful to the provided context and question'. Our problem is therefore how to define the concept of a 'most faithful' LLM response within our black box setting.  

\subsection{LLM as a Bipartite Information Engine: the LLM Maxwell's Demon}

We model an LLM as an {\it information engine} that consumes some input information (context $ C $ and user question $ Q $), transforms it, and produces its output $ A $. Furthermore, this information engine has a {\it bipartite structure}: it is made of two interacting stochastic sub-systems $ X $ and $ Y $ (thought of as a 'tape' and 'controller', respectively), that operate in consecutive steps. The first system $ X $ (the 'tape') is a part of the LLM exposed to the end user via a LLM user interface, roughly identified with the first and last layers of a neural network that implements the LLM. In particular, context $ C $ and the LLM answer $ A $ can be viewed as (noisy) observations of sub-system $ X $. This is achieved using a pre-trained sentence embedding model, and treating the sentence embedding vectors of $ C $ and $ A $ as noisy transforms of internal activations of the LLM in the process of answer generation. 

In contrast, the second sub-system $ Y $ (the controller) is {\it not} exposed to the user. This sub-system {\it controls} the transformation on the input context $ C $ into answer $ A $, as specified by the user question (prompt) $ Q $. 
This sub-system conceptualizes the notion of {\bf Maxwell's demon} as an agent that converts information into work.\footnote{See e.g. \cite{Parrondo_2015} on the history of the concept of Maxwell demon since James Clerk Maxwell's work to modern-day front lines of stochastic thermodynamics.} 
Sub-system $ Y $ is roughly identified with inner layers of the neural network.
While we do not observe states of sub-system $ Y $ either directly or indirectly, they should depend on embeddings of context $ C $ and question $ Q $, to the extend that the latter define the task for the LLM. 

The bipartite $ XY $-system proceeds in steps. First, the controller $ Y $ (the Maxwell demon) sets itself in a proper state by reading the user question $ Q $ and the context documents $ C $. This informs the LLM about a transformation of $ C $ needed to provide answer $ A $ expected by the user. The controller then defines a policy for the text generation, which is further used as a control protocol for the evolution of sub-system $ X $ from the initial state $ C $ to the final state $ A $.
Importantly, the two sub-systems $ X $ and $ Y $ of a bipartite information engine do not operate concurrently, but rather sequentially. This feature of bipartite systems make them amenable to theoretical analyses, see e.g. \cite{Ehrich_2023} for a review.  

The bipartite $ XY $-system representing the LLM can be analyzed using methods of both information theory and thermodynamics. We start with the information-theoretic analysis, and then present a complimentary view based on thermodynamics.    


\subsection{Topical Representation}

We view each of $\mathcal{Q}$, $\mathcal{C}$, and $\mathcal{A}$ as a collection of sentences. By analyzing the semantic embeddings of these sentences (e.g., using a pre-trained sentence transformer), we can identify a set of $N$ latent topics that span the semantic space of the triplet. We can then represent $\mathcal{Q}$, $\mathcal{C}$, and $\mathcal{A}$ as probability distributions over these $N$ topics. Let these marginal distributions be:
\begin{itemize}
    \item $p^{(c)} = (p_1^{(c)}, \dots, p_N^{(c)})$, the topic distribution of the context.
    \item $p^{(q)} = (p_1^{(q)}, \dots, p_N^{(q)})$, the topic distribution of the question.
    \item $p^{(a)} = (p_1^{(a)}, \dots, p_N^{(a)})$, the topic distribution of the answer.
\end{itemize}
We assume that these marginal distributions can be empirically estimated from the text triplet. In particular, they can be easily computed using the Semantic Divergence Metrics (SDM) method recently developed in  \cite{halperin2025sdm, halperin2025dib}. The SDM approach performs a joint clustering of sentence embeddings for all sentences in $ QCA $-triplets, and then identifies cluster topics by doing the term frequency analysis. Once clusters are found, all texts in the $ QCA $ triplet are converted into marginal probability distributions 
$ p^{(q)}, p^{(c)}, p^{(a)} $ by counting frequencies of cluster assignments of their sentences. 
 
We note here that while the SDM approach is assumed here as a practical and numerically inexpensive way to compute the marginal 
distributions $ p^{(q)}, p^{(c)}, p^{(a)} $, the method developed in this paper is agnostic to how these quantities are computed, and therefore in principle can be used alongside any other method that computes these marginal probabilities.     

\subsection{Information Flows as Topic Weights Transitions}

To model the flow of information from the context to the answer and from the context to the question, we introduce two $N \times N$ transition matrices, $\matQ = [Q_{ij}]$ and  $\matA = [A_{ij}]$, where:
\begin{itemize}
    \item $Q_{ij}$ is the conditional probability that a unit of topical probability mass from topic $i$ in the context is mapped to topic $j$ in the question.
    \item $A_{ij}$ is the conditional probability that a unit of topical probability mass from topic $i$ in the context is mapped to topic $j$ in the answer.
\end{itemize}
For these matrices to be valid, they must be row-stochastic, meaning the rows sum to one:
\begin{equation}
\label{normalization}
\sum_{j=1}^{N} Q_{ij} = 1 \quad \text{and}  \quad \sum_{j=1}^{N} A_{ij} = 1 \quad , \forall i \in \{1, \dots, N\}.
\end{equation}
Furthermore, these matrices must explain the observed marginal distributions. By the law of total probability, they must satisfy the following linear constraints:
\begin{align}
p_{j}^{(a)} &= \sum_{i=1}^{N} p_i^{(c)} A_{ij} , \quad \forall j \in \{1, \dots, N\} \label{eq:constraint_a} \\
p_{j}^{(q)}  &= \sum_{i=1}^{N} p_i^{(c)} Q_{ij}, \quad \forall j \in \{1, \dots, N\} \label{eq:constraint_q}
\end{align}
Note that the first equation here can be interpreted as a (one-step) Markov chain process with transition matrix $ {\bf A} $, which describes the topic evolution from the context $ C $ to the answer $ A $ produced by the LLM. The transition process can be thought of as proceeding in a time step $ \Delta T_a $.
On the other hand, we can similarly interpret the second equation in (\ref{eq:constraint_q}) as 
a dynamic transition of the initial topic distribution into a new distribution according to the prompt. This transition occurs in a shorter time step $ \Delta T_q \ll \Delta T_a $. Information in the user question (prompt) defines the objective (the desired 'semantic drift') for the task of transformation of the context text into the answer text by the LLM. 
Viewed from the point of view of the LLM as a bipartite $ XY$-system, the $ {\bf Q} $-dynamics are associated with an initial preparation of sub-system $ Y $, while sub-system $ X $
evolves according to the $ {\bf A}$-dynamics. When the condition  $ \Delta T_q \ll \Delta T_a $ holds, it means that sub-system  $ X $ remains idle for 
a very short time  $ \Delta T_q $ at the start of its response time period $  \Delta T_a $.   

We can expect that
matrices $ {\bf Q} $ and $ {\bf A} $ expressing, respectively, the 'objective' and  the 'result' of the topic transformation of the initial context $ C $ should be similar to each other.
Now consider our setting described above in Sect.~\ref{sect_problem}, 
where we have a set of $ K $ triplets
 $ (Q_i, C, A_i) $ with $ i = 1, \ldots, K $ enumerating the number of semantically equivalent prompt paraphrases. 
If we had a metric (score) to compare these triplets, we would be able to pick a triplet with the highest score as the most faithful one.  
As we will show in the next section, we can convert this idea into a lightweight numerical algorithm computing the faithfulness score for LLM answers.

\subsection{Semantic Faithfulness Score}

The constraints in Eqs.\eqref{normalization}-\eqref{eq:constraint_q} do not uniquely determine the matrices $\matA$ and $\matQ$. There can be many transition dynamics that satisfy the marginals. The key idea of our method is to choose the pair of matrices $(\matA, \matQ)$ that are minimally divergent from each other, reflecting the most "parsimonious" explanation for the transformations. We quantify this divergence using the conditional Kullback-Leibler (KL) divergence:
\begin{equation}
\label{KL}
D(\matA \parallel \matQ) = \sum_{i=1}^{N} p_i^{(c)} \sum_{j=1}^{N} A_{ij} \log \frac{A_{ij}}{Q_{ij}}.
\end{equation}
This objective measures the expected information-theoretic distance between the transition dynamics, where the expectation is taken over the starting topics as defined by the context distribution $p^{(c)}$.
The KL divergence (\ref{KL}) quantifies the information cost of encoding the transition matrix ${\bf A} $ in terms of transition matrix $ {\bf Q} $ \cite{Cover_IT_book}. As in our framework it is the Maxwell demon $ Y $ that eventually controls both matrices
$ {\bf Q} $ and ${\bf A} $, this information cost is carried by the Maxwell demon. 

Our proposed faithfulness  metric, which we call the \textit{Semantic Faithfulness} (SF) score and denote as $ \mathcal{F}_S $, is defined in terms of the minimal value of this KL divergence (\ref{KL}), obtained by jointly optimizing over $\matA$ and $\matQ$ subject to all constraints:
\begin{equation}
\label{F_S}
 \mathcal{F}_S := \frac{1}{1 + D_{min}}. \; \; \; D_{min} := \min_{\matA \in \mathcal{C}_\matA, \matQ \in \mathcal{C}_\matQ } D(\matA \parallel \matQ)
\end{equation}
Here  $ \mathcal{C}_{\matA} $ and $ \mathcal{C}_{\matQ} $ denote constraint sets for transition matrices $ {\bf A} $ and $ {\bf{Q}} $, respectively, according to Eqs.\eqref{normalization}-\eqref{eq:constraint_q}. As $ D_{min} $ can take values between zero and infinity, the 
SF score $ \mathcal{C}_{\matA} $ ranges from zero to one. Values of one are attained when the
two probability distributions $ {\bf A} $ and ${\bf Q} $ coincide.  This implies high faithfulness: the answer produces the same topical shifts that was implied by the question with respect to the context. Conversely, a small score suggests the answer's topical focus diverges significantly from the question's, indicating a potential lack of faithfulness. 

We hasten to stress here that the minimal KL divergence $  D_{min} $
defined in Eq.(\ref{F_S}) is a well-defined and unique quantity.
This follows
from the results of Csiszar and Tusn{\'a}dy \cite{csiszar1984}, who showed that the problem of a joint minimization of KL divergence 
$ D(\matA \parallel \matQ) $ has a solution as long as sets 
$ \mathcal{C}_{\matA} $ and $ \mathcal{C}_{\matQ} $ are convex, see also \cite{csiszar2004}. Furthermore, they also provided a constructive approach to find the solution by alternating minimization (AM) with respect to distributions 
$ {\bf A} $ and $ {\bf Q} $. 
When applied to the problem of joint minimization of $ D(\matA \parallel \matQ) $ under convex constraints, the AM method 
of Csiszar and Tusn{\'a}dy is known as the Blahut-Arimoto (BA) algorithm \cite{Blahut_1972, Arimoto_1972}. Convergence of the BA algorithm to a global minimum follows from joint convexity of $ D(\matA \parallel \matQ) $ in both its arguments, see \cite{Cover_IT_book}, p.30.
In the following section, we will present details of an AM scheme for our setting.

\section{Computing Semantic Faithfulness (Think Like the Demon!)}
\label{sect_AM_algorithm}

This section details the computational algorithm for finding the optimal matrices $\matA$ and $\matQ$ and thus computing the faithfulness  score.

\subsection{The Alternating Minimization (AM) Algorithm}

The overall optimization problem is to find the pair of matrices $(\matA, \matQ)$ that minimizes the objective function $D(\matA \parallel \matQ) $ subject to their respective constraint sets, which we denote $\mathcal{C}_\matA$ and $\mathcal{C}_\matQ$. 
As mentioned above, this problem is jointly convex in $\matA$ and 
$\matQ$, which provides convergence of the AM algorithm to a global minimum irrespective of a starting point. 

The Alternating Minimization (AM) algorithm 
\cite{csiszar1984, csiszar2004, Cover_IT_book} decomposes the joint optimization into a sequence of simpler sub-problems. Starting with an initial guess $\matQ^{(0)}$ for one matrix, the algorithm proceeds iteratively:
\begin{enumerate}
    \item \textbf{A-update Step:} With $\matQ^{(k)}$ held constant, solve for $\matA^{(k+1)}$:
        \[ \matA^{(k+1)} = \arg\min_{\matA \in \mathcal{C}_\matA} D(\matA \parallel \matQ^{(k)} ) \]
    \item \textbf{Q-update Step:} With the newly computed $\matA^{(k+1)}$ held constant, solve for $\matQ^{(k+1)}$:
        \[ \matQ^{(k+1)} = \arg\min_{\matQ \in \mathcal{C}_\matQ} D(\matA^{(k+1)} \parallel \matQ) \]
\end{enumerate}
This process is repeated until a convergence criterion is met.
We next describe each step in this procedure to derive updating
rules for matrices $\matA$ and $\matQ$.



\subsection{A-update Step}

For the A-update step, the objective is to minimize $D(\matA \parallel \matQ)$ for a fixed $\matQ$, subject to the constraints defining $\mathcal{C}_\matA$. The Lagrangian for this subproblem is:
\begin{equation}
\label{Lagrangian_step_1}
\mathcal{L}_1 (\matA, \boldlambda, \boldmu) = \sum_{i,j=1}^{N} p_i^{(c)} A_{ij} \log \frac{A_{ij}}{Q_{ij}} + \sum_{j} \lambda_j \left( \sum_{i=1}^{N} p_i^{(c)} A_{ij} - p_j^{(a)} \right) + \sum_{i=1}^{N} \mu_i p_i^{(c)}\left( \sum_{j} A_{ij} - 1 \right)
\end{equation}
where $\lambda_j$ and $\mu_i $ are Lagrange multipliers for the one-step evolution and row-stochasticity constraints, respectively. Here we additionally scaled the second (yet unknown) Lagrange multiplier
$ \mu_i $ by the factor
 $p_i^{(c)} $ in order to simplify formulas to follow. 
 
Setting the partial derivative of the Lagrangian with respect to $A_{ij}$ to zero produces the solution for $A_{ij}$:
%
\begin{equation}
\label{A_sol}
A_{ij} = Q_{ij} e^{- \lambda_j - \mu_i - 1} 
= \frac{ Q_{ij} e^{- \lambda_j }}{\sum_{j=1}^{N} 
Q_{ij} e^{- \lambda_j}}
\end{equation}
Here at the last step we eliminated the normalization Lagrange 
multipliers $ \mu_i $ by enforcing the normalization constraint 
$ \sum_{j} {\bf A}_{ij} = 1 $. 
To find the Lagrange multipliers $ \lambda_j $, we plug the solution back to Eq.(\ref{Lagrangian_step_1}) to obtain the dual Lagrangian
that should be maximized with respect to $ \lambda_j $:
\begin{equation}
\label{Lagrangian_step_1_dual}
\mathcal{L}_1 (\lambda) = \sum_{i,j} p_i^{(c)} 
\log \left( \sum_{j=1}^{N} {\bf Q}_{ij} e^{- \lambda_j} \right)
- \sum_{j=1}^{N} \lambda_j p_j^{(a)}
\end{equation}
Equating the partial derivatives of this Lagrangian to zeros, we obtain equations for $ \lambda_j $:
\begin{equation}
e^{- \lambda_j} \sum_{i=1}^{N}  \frac{ p_i^{(c)} {\bf Q}_{ij}}{
\sum_{j=1}^{N} {\bf Q}_{ij} e^{- \lambda_j} } = p_j^{(a)}
\end{equation}
Instead of relying on numerical convex optimization, these equations can be solved semi-analytically by converting them to fixed-point point equations for expressions 
$ u_j := e^{- \lambda_j} $:
\begin{equation}
\label{fixed_point}
 u_j = \frac{ p_j^{(a)}}{ \sum_{i=1}^{N}  \frac{ p_i^{(c)} {\bf Q}_{ij} }{
\sum_{j=1}^{N} {\bf Q}_{ij} u_j }}, \; \;  u_j := e^{- \lambda_j}
\end{equation}
and solving them by iterations. Once the fixed point $ {\bf u} = {\bf u}_{\star} $ is found, we obtain matrix $ {\bf A} $ using 
Eq.(\ref{A_sol}) as follows:
\begin{equation}
\label{A_sol_u}
A_{ij} 
= \frac{ Q_{ij} u_j}{\sum_{j=1}^{N} 
Q_{ij} u_j}
\end{equation}
Therefore, the $A$-update step of our algorithm can be done semi-analytically.

\subsection{Q-update Step}

For the $Q$-update step, the objective is to minimize $D(\matA \parallel \matQ)$ for a fixed $\matQ$, subject to the constraints defining $\mathcal{C}_\matQ$. The Lagrangian for this sub-problem is:
\begin{equation}
\label{Lagrangian_step_2}
\mathcal{L}_2 (\matQ, \boldxi, \boldmu) = \sum_{i,j=1}^{N} p_i^{(c)} A_{ij} \log \frac{A_{ij}}{Q_{ij}} + \sum_{j=1}^{N} \xi_j \left( \sum_{i} p_i^{(c)} Q_{ij} - p_j^{(q)} \right) + \sum_{i=1}^{N} \nu_i p_i^{(c)}\left( \sum_{j} Q_{ij} - 1 \right)
\end{equation}
where $\xi_j$ and $\nu_i $ are the Lagrange multipliers for the one-step evolution and row-stochasticity constraints, respectively.
Setting the derivative $\frac{\partial \mathcal{L}_2}{\partial Q_{ij}} = 0$ gives:
\[ - \frac{A_{ij}}{Q_{ij}} + \nu_i + \xi_j = 0 \]
Solving for $Q_{ij}$, we obtain
\begin{equation}
Q_{ij} = \frac{A_{ij}}{\nu_i + \xi_j }
\end{equation}
Plugging this solution back to Eq.(\ref{Lagrangian_step_2}), we obtain 
the dual Lagrangian that should be maximized with respect to  Lagrange multipliers $ \boldxi, \boldnu $:
\begin{equation}
\label{Lagrangian_step_2_dual}
\mathcal{L}_2 (\boldxi, \boldnu) = \sum_{i,j=1}^{N} p_i^{(c)} A_{ij} 
\log \left( \nu_i + \xi_j \right)  - \sum_{i=1}^{N} p_i^{(c)} \nu_i - 
  \sum_{j=1}^{N} p_j^{(q)} \xi_j + 1
\end{equation}
It is easy to verify that thus Lagrangian is separately concave in $ \boldnu $ when $ \boldxi $ fixed, and in  $ \boldxi $ when  $ \boldnu $ is fixed. Therefore, it can be quickly maximized using alternating maximization with respect to  $ \boldnu $ and $ \boldxi $ while keeping the other parameter fixed.

The $ Q$-step and $ A $-step described here are iterated until convergence. The final Semantic Faithfulness calculation is summarized in Algorithm \ref{SF_algorithm}.

\begin{algorithm}[h!]
\caption{The Semantic Faithfulness Algorithm}
\label{SF_algorithm}
\begin{algorithmic}[1]
\State \textbf{Input:} Marginal distributions $p^{(c)}$, $p^{(a)}$,  $p^{(q)}$, initial matrix $\matQ^{(0)}$, tolerances $\epsilon_{\text{outer}}, \epsilon_{\text{inner}}$.
\State \textbf{Initialize:} Set outer loop counter $k=0$.
\Repeat
    \Statex \textit{// A-Step (Projection onto $\mathcal{C}_\matA$)}
    \State Given $\matQ^{(k)}$, find $\matA^{(k+1)}$ in two steps:
    \State Find scaling factors $u_j$ for all $j$ by solving 
    Eq.(\ref{fixed_point})
    \State Compute matrix  $\matA^{(k+1)}$ using Eq.(\ref{A_sol_u})

    \Statex \textit{// Q-Step (Projection onto $\mathcal{C}_\matQ$)}
    \State Given $\matA^{(k+1)}$, find $\matQ^{(k+1)}$ in two steps:
    \State Compute Lagrange multipliers $\boldxi $ and $ \boldnu$ by alternating maximization:
    \Repeat
        \State  Maximize the Lagrangian (\ref{Lagrangian_step_2_dual}) for $ \boldnu $ with keeping $ \boldxi $ fixed.
        \State Maximize the Lagrangian (\ref{Lagrangian_step_2_dual}) for  $ \boldxi $ with keeping  $ \boldnu $ fixed.
    \Until{changes in $\boldnu, \boldxi$ are less than $\epsilon_{\text{inner}}$}
    \State Compute the updated matrix: $Q_{ij}^{(k+1)} = \frac{ A_{ij}^{(k+1)}}{\nu_i + \xi_j}$.
    \Statex \textit{// Check Convergence}
    \State Increment $k \leftarrow k+1$.
\Until{change in objective function $D(\matA^{(k)} \parallel \matQ^{(k)}; p^{(c)})$ is less than $\epsilon_{\text{outer}}$}
\State \textbf{Output:} Converged matrices $\matA^*, \matQ^*$, the minimal divergence $D_{min} = D(\matA^* \parallel \matQ^*)$, the SF score $ 
\mathcal{F}_S = 1/(1+D_{min}) $.
\end{algorithmic}
\end{algorithm}

\section{Semantic Entropy Production}
\label{sect_Entropy_Production}

So far we developed a semantic model for the process of text generation by a LLM as stochastic dynamics of transitions from the input context to the LLM answer, defined on a discrete-valued semantic (topic) space, and modulated by the user question that serves as a control variable. 

Our analysis so far was only based on information-theoretic methods. Here we present a complimentary view of the same dynamics of the LLM as a bipartite $ XY$-information engine, this time analyzed as a {\it physical engine}. More specifically, we want to address {\it thermodynamics} of our bipartite information engine.  
For a review of thermodynamics of information, see e.g. \cite{Parrondo_2015}.

To recall the setting of our bipartite $ X Y $ model, both stochastic sub-systems $ X $ and $ Y $ (the tape and controller, respectively) interact with each other and with their respective heat baths. We now want to explore transformations in sub-system $ X $, i.e. transitions from context $ C $ to answer $ A $ controlled by prompt $ Q $, from the point of view of thermodynamics.  

In general, such transitions proceed out of equilibrium, due to both a non-equilibrium starting point (the context) and applied control by the prompt. Such non-equilibrium transitions are accompanied by {\bf entropy production}. Entropy production quantifies the amount of time non-reversibility in a 
given (non-equilibrium) process, and thus defines the direction of the arrow of time, see e.g. \cite{Spinney_2012} for a review.    

A part of entropy produced during such transitions is dissipated as heat into the environment of the system. If the environment needs to be kept at a fixed temperature, this implies that larger entropy production generally increases costs of cooling the system to maintain a fixed bath temperature. In its turn, this means that an optimal control should minimize entropy production. 

In classical statistical mechanics, the concept of entropy production is normally defined at the level of a statistical {\it ensemble} of many trajectories of a statistical system with $ N \rightarrow \infty $ particles\footnote{More precisely, classical thermodynamics deals with macroscopic systems with $N \sim N_A $ particles, where $ N_A \simeq 6.02 \times 10^{23} $ is the Avogadro constant.}, sampled from a given stochastic process.
In contrast, the modern field of {\it stochastic thermodynamics}
studies much smaller ('mesoscopic') systems 
where systems and ensembles can be really small, with $ N \sim 10 - 10^{3} $. Furthermore, in stochastic thermodynamics, the concept of entropy production is defined at the level of both {\it individual trajectories} and ensembles of trajectories 
\cite{Seifert_2012}.  
The ability to compute entropy production in such mesoscopic systems  at the level of single trajectories offers a possibility to monitor and control individual trajectories by choosing control protocols that minimize entropy production along these trajectories.

In our setting of of a user-LLM interaction with a given context $ C $, a dataset of $ K $ triplets $ (Q_k, C, A_k) $ with $ k = 1, \ldots, K $ can be viewed as a set of 
triplets of transitions of sub-system $ X $ between states $ C \rightarrow A_k $ produced by 'actions' $ Q_k $ which effect the stage of controller $ Y $. Importantly, we want to analyze and compute entropy production for each triplet separately in order to be able to compare them. On the other hand, each such triplet is itself 
a small ensemble over topics contained in 
$ (Q_k, C, A_k) $, as we do not observe individual 'topic trajectories' that would start with a single topic, rather than the whole prompt-context combination $ Q, C $.


This view of the LLM enables employing methods of computing entropy production from stochastic thermodynamics in our problem of comparison of individual $ QCA $-triplets. 
The prompt-answer pair that gives rise to the minimum entropy production may be suggested as the best (most efficient, least heat-dissipating) candidate in the set. 

As shown in \cite{Seifert_2005}, total entropy production for a stochastic system is given by the KL divergence between the probabilities of the forward and backward (time-reversed) paths.
In our one-step setting, this produces the following expression for the total entropy production $ \Sdot_{tot} $:
\bea
\label{Sdot_tot}
\Sdot_{tot} &=& \sum_{i,j=1}^{N} p_{i}^{(c)} A_{ij} \log \frac{ p_i^{(c)} 
A_{ij}}{ p_j^{(a)} A_{ji}^{R} } = 
\sum_{i,j=1}^{N} p_{i}^{(c)} A_{ij} \log \frac{ 
A_{ij}}{A_{ji}^{R} } + \sum_{i,j=1}^{N}  p_{i}^{(c)} A_{ij} \log \frac{ p_i^{(c)}}{ p_j^{(a)} } \nonumber \\
&=& \sum_{i,j=1}^{N} p_{i}^{(c)} A_{ij} \log \frac{ 
A_{ij}}{A_{ji}^{R} } + H \left[ p^{(a)} \right] - H \left[ p^{(c)} \right]
\eea     
Here $ {\bf A}^R $ stands for the transition matrix of the time-reversed process. As this quantity is not directly measured from our data, Eq.(\ref{Sdot_tot}) is not sufficient on its own for computing entropy production without additional assumptions or approximations.
Before we proceed with our estimation of this quantity, we pause to make a few remarks.

\subsection{Decomposition of the Total Entropy Production}

As discussed in \cite{Seifert_2005}, the structure of Eq.(\ref{Sdot_tot}) suggests the following decomposition for the total entropy production:
\beq
\label{EP_decomp}
\Sdot_{tot} =  \Sdot_{m} + \Sdot 
\eeq
where
\beq
\label{Sdot_m}
\Sdot_{m} = \sum_{i,j=1}^{N} p_{i}^{(c)} A_{ij} \log \frac{ 
A_{ij}}{A_{ji}^{R} }
\eeq 
is the entropy (or, equivalently, heat) dissipated to the heat bath (medium), and 
\beq
\label{S_dot}
\Sdot = \sum_{i,j=1}^{N}  p_{i}^{(c)} A_{ij} \log \frac{ p_i^{(c)}}{ p_j^{(a)} } =  H \left[ p^{(a)} \right] - H \left[ p^{(c)} \right]
\eeq
is the entropy change of sub-system $ X $. We next separately discuss these two contributions to the total entropy production $ \Sdot_{tot} $ in sub-system $ X $. 
 
\subsection{Semantic Entropy or Semantic Entropy Production?}
\label{sect_SE_or_SEP}

Let us consider first the sub-system $ X $ entropy change term (\ref{S_dot}). It amounts to the difference $  H \left[ p^{(a)} \right] - H \left[ p^{(c)} \right] $ of marginal final and initial entropies. The first term here is the semantic Shannon entropy of the LLM answer. This quantity was previously suggested on heuristic grounds in the Semantic Entropy (SE) method \cite{farquhar2024semantic} as a metric for LLM hallucination control. As was noted in \cite{halperin2025sdm}, one drawback of the SE method is that it does not account for complexity of the initial context (or prompt). 

On the other hand, the present work suggests that it is entropy production, rather than the marginal entropy of the LLM answer, that may be a better metric for quantification of uncertainty and confidence of LLM answers. We see that the system entropy change
(\ref{S_dot}) appears a more meaningful way to quantify entropy of the LLM answer by computing its difference with entropy of the context. Furthermore, the concept of entropy production is intuitively related to the notion of LLM hallucinations as noisy (entropy-increasing) distortions of the input context and prompt data. 
  
Now, the sub-system $ X $ entropy change (\ref{S_dot}) can be directly calculated from the marginal distribution, but it amounts only to one contribution into the total entropy production according to Eq.(\ref{EP_decomp}). The other term $ \Sdot_{m} $ defined 
in Eq.(\ref{Sdot_m}) cannot be directly calculated as long as the reverse transition matrix $ {\bf A}^R $ is not known or estimated.
We will next propose a method to estimate the dissipated entropy
$ \Sdot_{m} $ by computing its lower bound.

\subsection{Lower Bound on Semantic Entropy Production}
\label{sect_lower_bound}  

As the reverse transition matrix $ {\bf A}^R $ is not directly measured, the exact amount of entropy production according to Eq.(\ref{Sdot_tot}) is unknown. 
However, we can find a {\it lower bound} on entropy production in our process by minimizing the expression (\ref{Sdot_tot}) with respect to all possible reverse transition matrices $ {\bf A}^R $, subject to all constraints that should be imposed on these matrices \cite{Ito_2020}.
In our case, matrix $ {\bf A}^R $ should satisfy the marginal and normalization constraints
\begin{equation}
\label{marginaL_A_R}
\sum_{j=1}^{N} p_j^{(a)} A_{ji}^{R} = p_i^{(c)}, \; \; \; 
\sum_{i=1}^{N} A_{ji}^{R} = 1
\end{equation}
 (\ref{marginaL_A_R}). This produces the following Lagrangian function: 
\begin{equation}
\label{Lagrangian_S}
\mathcal{L}_S ({\bf A}^R, \boldxi, \boldnu) = \sum_{i,j} p_i^{(c)} A_{ij} \log \frac{A_{ij}}{A_{ij}^R} + \sum_{i} \xi_i 
\left( \sum_{i} p_j^{(a)} A_{ji}^R - p_i^{(c)} \right) + \sum_{j} \nu_j p_j^{(c)}\left( \sum_{i} A_{ji}^R - 1 \right)
\end{equation}
where $\xi_j$ and $\nu_j $ are Lagrange multipliers. 
Minimization of this Lagrangian with respect to $ {\bf A}^R $ gives
\beq
\label{A_R_sol}
A_{ji}^R = \frac{p_i^{(c)}}{\xi_i p_j^{(a)} + \nu_j p_j ^{(c)}} 
A_{ij}
\eeq  
Plugging this back into (\ref{Lagrangian_S}), rescaling the optimization variables $ \nu_j \rightarrow \nu_j p_j^{(c)}/p_j^{(a)} $ and simplifying the resulting expression, we obtain the dual Lagrangian that should be maximized with respect to the Lagrange multipliers:
\bea
\label{L_S_dual}
\mathcal{L}_S (\boldxi, \boldnu) 
&=& \sum_{i,j} p_i^{(c)} A_{ij} \log 
\left( \xi_i + \nu_j \right) - \sum_{i} p_i^{(c)} \xi_i - 
\sum_{j} p_j^{(a)} \nu_j + 1 \nonumber \\
&=& \mathcal{L}_2(\boldxi, \boldnu) + \sum_{j=1}^{N} \nu_j \left( 
p_j^{(q)} - p_j^{(a)} \right)
\eea 
where $ \mathcal{L}_2(\boldxi, \boldnu) $ is the Lagrangian defined in Eq.(\ref{Lagrangian_step_2_dual}). Therefore, the remaining maximization in Eq.(\ref{L_S_dual}) can be done using the same objective functions in in the $Q$-step of our Semantic Faithfulness algorithm, upon the substitution $ {\bf p}^{(q)} \rightarrow 
{\bf p}^{(a)} $. Alternatively, if the second term in the last expression in Eq.(\ref{L_S_dual}) is much smaller than the first term, the optimal value of $ \mathcal{L}_S (\boldxi, \boldnu) $
can be computed to the first order in the perturbation as follows:
\beq
\label{S_dual_first_order}
\mathcal{L}_S (\boldxi^{\star}, \boldnu^{\star}) \simeq
\mathcal{L}_2 (\boldxi^{\star}, \boldnu^{\star}) 
 + \sum_{j=1}^{N} \nu_j^{\star} \left( 
p_j^{(q)} - p_j^{(a)} \right)   
\eeq
where $ \boldxi^{\star}, \boldnu^{\star} $ are the optimal Lagrange multipliers for the Lagrangian $ \mathcal{L}_2 (\boldxi, \boldnu) $.
The final scheme is presented in Algorithm~\ref{alg:sep}.

\begin{algorithm}[H]
  \caption{Semantic Entropy Production (SEP)}
  \label{alg:sep}
  \begin{algorithmic}[1]
  \Require $\mathbf{A}^*$ (optimal transition matrix from Algorithm~\ref{SF_algorithm}), $\mathbf{p}^{(c)}$, $\mathbf{p}^{(a)}$
  \State Initialize $\boldsymbol{\xi}, \boldsymbol{\nu} \in \mathbb{R}^N_{>0}$
  \Repeat
      \State \textbf{$\boldsymbol{\xi}$-step:} Maximize $\mathcal{L}_S(\boldsymbol{\xi}, \boldsymbol{\nu})$ w.r.t. $\boldsymbol{\xi}$ (fixing $\boldsymbol{\nu}$):
      \begin{equation}
      \xi_i^* = \argmax_{\xi_i} \left[ \sum_j p_i^{(c)} A^*_{ij} \log(\xi_i + \nu_j) - p_i^{(c)} \xi_i \right]
      \end{equation}
      \State \textbf{$\boldsymbol{\nu}$-step:} Maximize $\mathcal{L}_S(\boldsymbol{\xi}, \boldsymbol{\nu})$ w.r.t. $\boldsymbol{\nu}$ (fixing $\boldsymbol{\xi}$):
      \begin{equation}
      \nu_j^* = \argmax_{\nu_j} \left[ \sum_i p_i^{(c)} A^*_{ij} \log(\xi_i + \nu_j) - p_j^{(a)} \nu_j \right]
      \end{equation}
  \Until{convergence of $\mathcal{L}_S(\boldsymbol{\xi}, \boldsymbol{\nu})$}
  \State Recover $A^R_{ji} = \dfrac{A^*_{ij}}{\xi_i^* + \nu_j^*}$
  \State \Return SEP $= D(\mathbf{A}^* \| \mathbf{A}^R) = \sum_{i,j} p_i^{(c)} A^*_{ij} \log \dfrac{A^*_{ij}}{A^R_{ji}}$
  \end{algorithmic}
  \end{algorithm}

%
%
%
%
%
%

\subsection{Relationship Between Faithfulness and Entropy Production}
  \label{sect_Faithfulness_and_EP}

  The approximate expression (\ref{S_dual_first_order}) suggests
  a link between the (lower bound of) entropy production and
  the Faithfulness Score (\ref{F_S}). As the first term in (\ref{S_dual_first_order}) is exactly the quantity $ D_{min} =  1/
  \mathcal{F}_S - 1 $ (with added constraint penalties), Eq.(\ref{S_dual_first_order}) can be written as follows:
  \beq
  \label{Sdot_F_S}
  \Sdot_{tot} = 1/\mathcal{F}_S - 1  + \sum_{j=1}^{N} \nu_j^{\star} \left(
  p_j^{(q)} - p_j^{(a)} \right)
  \eeq
  When the correction term (the sum) is negligible, this yields the ``naive'' approximation:
  \beq
  \label{Sdot_F_S_naive}
  \Sdot_{tot} \approx D_{min} = \frac{1}{\mathcal{F}_S} - 1
  \eeq

\begin{figure}[H]
  \centering
  \includegraphics[width=0.6\textwidth]{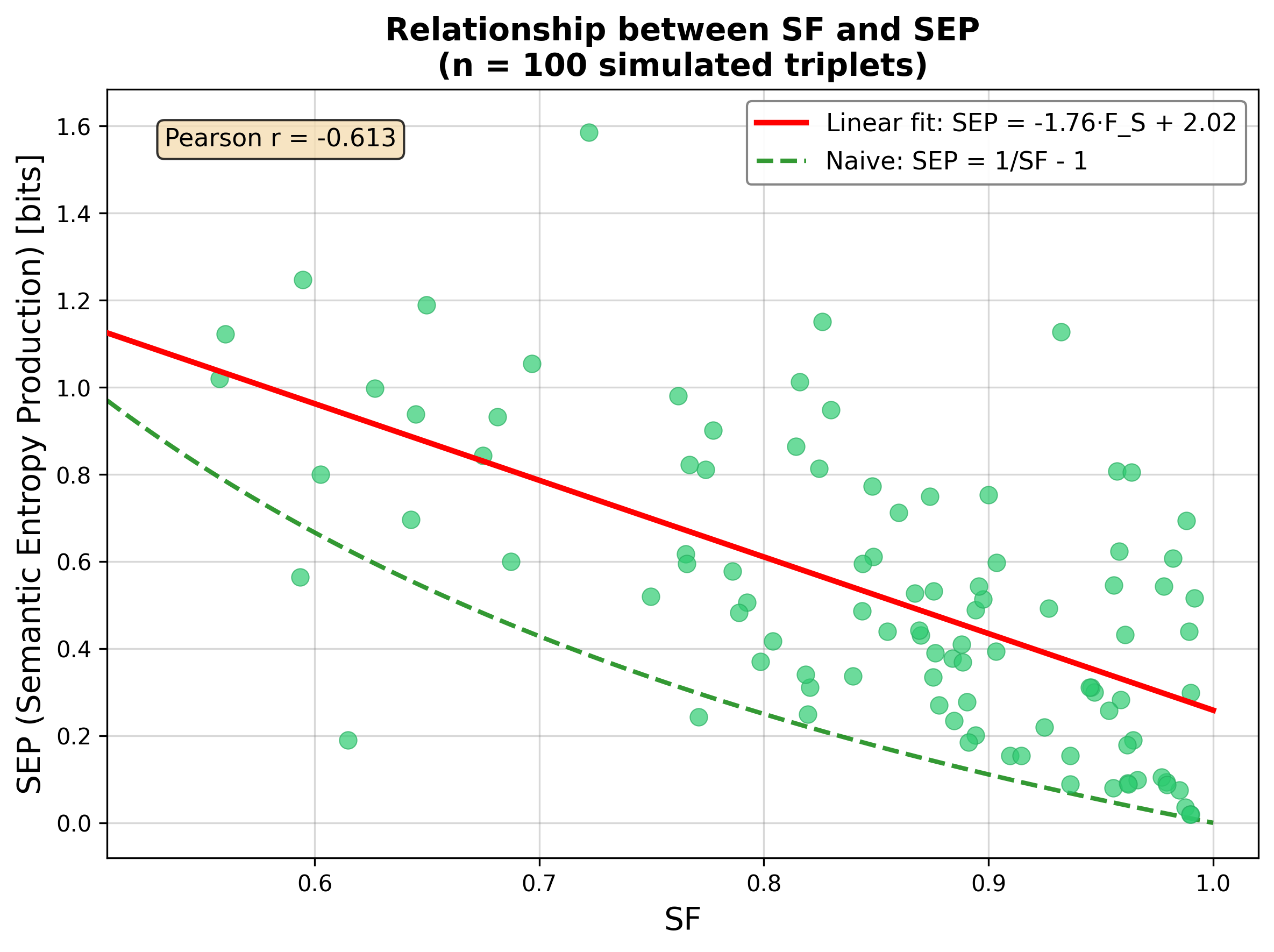}
  \caption{Scatter plot of Semantic Faithfulness ($\mathcal{F}_S$) versus Semantic Entropy Production (SEP) for $n=100$ simulated QCA triplets. The solid red line shows the linear
  regression fit (SEP $= -1.76 \cdot \mathcal{F}_S + 2.02$), while the dashed green line shows the naive approximation SEP $= 1/\mathcal{F}_S - 1$ from Eq.~(\ref{Sdot_F_S_naive}).}
  \label{fig:sf_sep_scatter}
  \end{figure}

To explore the relationship between $\mathcal{F}_S$ and SEP beyond such naive approximation, we generated 100 synthetic QCA triplets that preserve the key statistical properties observed in the real data presented in next section. Specifically, we sampled distributions $p(Q)$, $p(C)$, and $p(A)$ using Dirichlet distributions with concentration parameters
 calibrated to match the sparsity patterns that we observe in real datasets: questions exhibit high sparsity (concentrated on few topics), contexts are more diffuse (spread across many topics),
  and answers show intermediate sparsity.
We also preserved the empirical co-dependencies by sampling answer distributions conditionally on context distributions, reflecting the fact that LLM answers draw selectively from the provided context. 
For each synthetic triplet, we computed both the SF ($\mathcal{F}_S$), and SEP metrics using, respectively, Algorithms 1 and 2.
Figure~\ref{fig:sf_sep_scatter} presents a scatter plot of SF versus SEP, comparing the naive approximation (\ref{Sdot_F_S_naive}) with the exact calculation. 

This analysis of simulated triplets reveals a moderate negative correlation between $\mathcal{F}_S$ and SEP (Pearson $r = -0.61$), confirming that higher semantic faithfulness generally corresponds to lower entropy production. However, the relationship is weaker than the naive approximation SEP $\approx 1/\mathcal{F}_S - 1$ would
  suggest. A linear fit yields SEP $= -1.76 \cdot \mathcal{F}_S + 2.02$, indicating that while the metrics are related, they capture distinct aspects of information flow in QCA triplets, and should therefore be
computed independently, without reliance on the naive approximation 
(\ref{Sdot_F_S_naive}).

\subsection{Final Metrics: Semantic Faithfulness and Semantic Entropy Production}

  Our framework provides two complementary metrics for LLM faithfulness evaluation:
  \begin{enumerate}
  \item \textbf{Semantic Faithfulness (SF)}: The score $\mathcal{F}_S = 1/(1+D_{min})$ defined in Eq.~(\ref{F_S}), computed via the alternating minimization
  Algorithm~\ref{SF_algorithm}. This metric quantifies the information-theoretic alignment between the question-induced and answer-induced topic transformations.

  \item \textbf{Semantic Entropy Production (SEP)}: The lower bound on total entropy production $\Sdot_{tot}$ computed via a separate optimization as described in
  Section~\ref{sect_lower_bound}. This metric quantifies the thermodynamic irreversibility of the answer generation process.
  \end{enumerate}

\begin{figure}[H]
  \centering
  \includegraphics[width=0.9\textwidth]{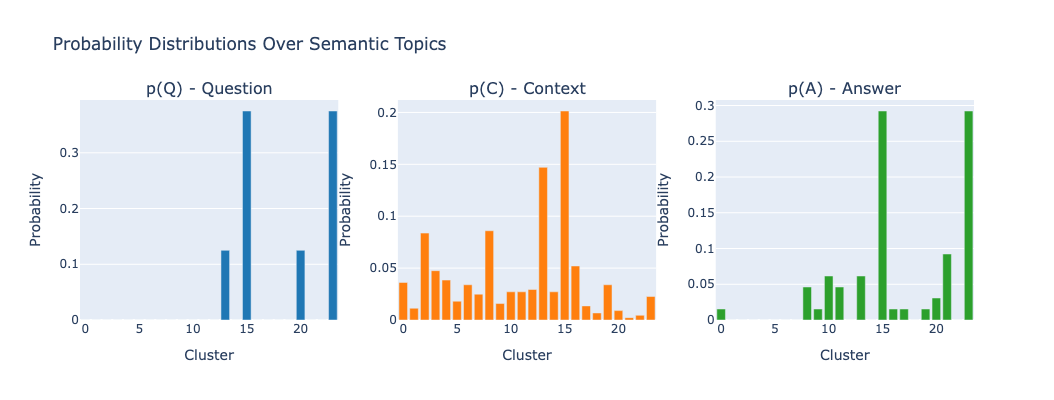}
  \caption{Probability distributions over semantic topics for triplet A0. Left: Question distribution $p(Q)$ is sparse, concentrated on a few semantic clusters. Center: Context
  distribution $p(C)$ is more diffuse, covering many topics from the source document. Right: Answer distribution $p(A)$ shows intermediate sparsity, reflecting how the LLM selectively
  addresses topics from the context to answer the question. These distributions serve as inputs to the SF and SEP algorithms.}
  \label{fig:marginal_distributions}
  \end{figure}

\begin{figure}[H]
  \centering
  \includegraphics[width=0.9\textwidth]{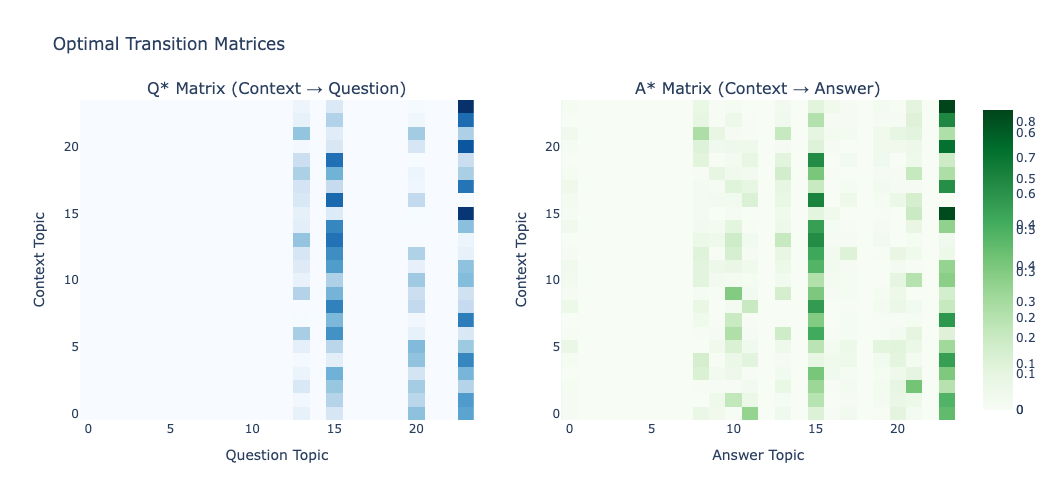}
  \caption{Optimal transition matrices $\mathbf{Q}^{\star}$ (left, blue colormap) and $\mathbf{A}^{\star}$ (right, green colormap) inferred by the alternating minimization algorithm for
   triplet A0. The $\mathbf{Q}^{\star}$ matrix encodes the optimal topic transformation from context to question, while $\mathbf{A}^{\star}$ encodes the transformation from context to
  answer. Both matrices exhibit sparse structure with probability mass concentrated along specific topic mappings. The Semantic Faithfulness score $\mathcal{F}_S$ quantifies how closely
   these two matrices align.}
  \label{fig:optimal_matrices}
  \end{figure}

Our simulation results in Figure~\ref{fig:sf_sep_scatter} demonstrate
that SF and SEP are related but not redundant, as they capture distinct, though connected, aspects of QCA triplet quality.
Indeed, SF is the information-theoretic metric that captures the \emph{semantic alignment} between question intent and answer content. On the other hand, 
SEP captures the \emph{thermodynamic efficiency} of the information transformation. We therefore recommend computing both metrics independently for comprehensive faithfulness evaluation. As a byproduct of the optimization algorithms, we also obtain the optimal
  transition matrices $\mathbf{Q}^{\star}$ and $\mathbf{A}^{\star}$ that achieve the minimal divergence, providing interpretable representations of topic flow from context to question
and answer, see Figures \ref{fig:marginal_distributions} and \ref{fig:optimal_matrices} below for typical marginal distributions and optimal transition matrices, respectively, that are obtained with our datasets. Details of our datasets will be presented in the next section.

\section{Experiments}
\label{sec:experiments}

We conducted experiments to validate our Semantic Faithfulness framework on a dataset of 10 rich Question-Context-Answer (QCA) triplets.\footnote{All code, data, and reproducibility materials are available at: \url{https://github.com/ighalp/semantic-faithfulness-sdm}} Each triplet consisted of a carefully crafted question, a comprehensive multi-paragraph context document, and a detailed LLM-generated answer.

\subsection{Dataset Construction}

\paragraph{Context Document.}
The context for all QCA triplets was extracted from NVIDIA Corporation's fiscal year 2024 Annual Report (Form 10-K), specifically the comprehensive Risk Factors section (Item 1A). This section spans approximately 60,000 characters and provides detailed disclosures of risks across multiple categories:
\begin{itemize}
    \item Risks related to industry and markets (competitive dynamics, evolving customer needs)
    \item Risks related to demand, supply, and manufacturing (supply chain dependencies, demand forecasting)
    \item Risks related to global operations (macroeconomic factors, international exposure, cybersecurity)
    \item Regulatory, legal, and governance risks (export controls, IP protection, tax policies)
\end{itemize}

\paragraph{Question Design Strategy.}
To systematically investigate the relationship between question semantic structure and Semantic Faithfulness, we constructed 10 QCA triplets organized into two groups based on semantic focus:

\begin{itemize}
    \item \textbf{Group A (Comprehensive Risk Analysis):} 5 paraphrases of a broad multi-topic question covering NVIDIA's entire risk landscape (supply chain, competition, product lifecycle, regulations, etc.). These paraphrases vary in phrasing and emphasis while maintaining the same comprehensive scope, resulting in moderate question entropy variation (CV $\approx$ 25\%).
    \item \textbf{Group B (Competitive Threats Focus):} 5 paraphrases of a qualitatively different question focused specifically on competitive dynamics (hyperscaler custom silicon, AMD/Intel alternatives, CUDA lock-in erosion, workload evolution). These paraphrases exhibit tight semantic clustering (CV $\approx$ 2\%), with all questions converging on the competitive threat theme.
\end{itemize}

\paragraph{Answer-Question Generation.}
All questions were answered using Claude Sonnet 4.5 (Anthropic), a state-of-the-art large language model. Each question shared the same context (NVIDIA 10-K Risk Factors disclosure, $\sim$60,000 characters). Group A questions elicited comprehensive multi-section risk analyses ($\sim$15,000 characters) covering all major risk categories, while Group B questions produced focused competitive analyses ($\sim$20,000 characters) with detailed technical comparisons of alternative AI accelerators and strategic implications.
All answers were generated using Gemini-2.5-Pro LLM model.

\paragraph{Computational Pipeline.}
For each QCA triplet, we computed semantic distributions over topics using the following efficient caching strategy:
\begin{enumerate}
    \item \textbf{Embedding:} All text was decomposed into sentences (1,514 total sentences across prompts, context, and answers) and embedded using the \texttt{Qwen3-Embedding-0.6B} model in a \emph{single} embedding pass, producing dense vector representations cached to disk.
    \item \textbf{Clustering:} The joint embedding space was clustered into $N=23$ semantic topics using the Upper-Bounded Deterministic Information Bottleneck (UDIB) algorithm~\cite{halperin2025dib} in a \emph{single} clustering pass. The UDIB method simultaneously performs clustering and determines the optimal number of clusters by maximizing the information bottleneck objective, discretizing the continuous semantic space into interpretable topic distributions.
    \item \textbf{Distributions:} For each triplet's question $Q$, context $C$, and answer $A$, we computed the probability distribution over topics by aggregating sentence cluster assignments and normalizing counts to obtain $p_q$, $p_c$, and $p_a$.
    \item \textbf{Caching:} All embeddings, cluster labels, and distributions were pre-computed once and cached, enabling instantaneous metric evaluation across multiple runs without redundant computation (saving $\sim$5-10 minutes per analysis).
\end{enumerate}

\subsection{Results}

  Table~\ref{tab:results} presents the computed metrics for all 10 QCA triplets. Context entropy $H(C) = 3.279$ bits is constant across all triplets as they share the same source
  document. All entropy values are reported in bits.

  \begin{table}[H]
  \centering
  \caption{Entropy values and computed metrics for all QCA triplets.}
  \label{tab:results}
  \begin{tabular}{llcccccc}
  \toprule
  Triplet & Group & $H(Q)$ & $H(C)$ & $H(A)$ & $\Sdot$ & $\mathcal{F}_S$ & SEP \\
  \midrule
  A0 & A & 1.447 & 3.279 & 3.905 & 0.627 & 0.472 & 0.771 \\
  A1 & A & 2.000 & 3.279 & 3.864 & 0.585 & 0.477 & 0.596 \\
  A2 & A & 3.122 & 3.279 & 4.107 & 0.829 & 0.577 & 0.076 \\
  A3 & A & 2.922 & 3.279 & 4.022 & 0.744 & 0.502 & 0.149 \\
  A4 & A & 2.500 & 3.279 & 4.129 & 0.851 & 0.516 & 0.202 \\
  \midrule
  B0 & B & 2.683 & 3.279 & 3.672 & 0.393 & 0.523 & 0.242 \\
  B1 & B & 2.585 & 3.279 & 3.666 & 0.387 & 0.477 & 0.280 \\
  B2 & B & 2.777 & 3.279 & 3.596 & 0.318 & 0.528 & 0.225 \\
  B3 & B & 2.689 & 3.279 & 3.664 & 0.386 & 0.489 & 0.103 \\
  B4 & B & 2.624 & 3.279 & 3.663 & 0.384 & 0.500 & 0.227 \\
  \midrule
  \multicolumn{2}{l}{Group A mean} & 2.398 & 3.279 & 4.006 & 0.727 & 0.509 & 0.359 \\
  \multicolumn{2}{l}{Group B mean} & 2.671 & 3.279 & 3.652 & 0.374 & 0.503 & 0.215 \\
  \multicolumn{2}{l}{Overall mean} & 2.535 & 3.279 & 3.829 & 0.550 & 0.506 & 0.287 \\
  \bottomrule
  \end{tabular}
  \end{table}

  \subsection{Analysis and Discussion}

  Our experimental results produced several important observations regarding the relationship between question structure, semantic faithfulness, and entropy production.

  \paragraph{Meaningful Question Entropy Variation.}
  Our question set achieves meaningful entropy variation. All 10 triplets exhibit substantial question entropy ($H(Q) \in [1.447, 3.122]$ bits, mean = 2.535 bits, CV = 18\%). The
  controlled two-group design demonstrates considerably different variance: Group B (competitive threats) shows tight clustering (CV = 2.4\%), while Group A (comprehensive risk)
  exhibits broader variation (CV = 25.5\%).

  \paragraph{Question Semantic Structure Impact.}
  Both groups exhibit similar mean Semantic Faithfulness (Group A: $\mathcal{F}_S = 0.509$, Group B: $\mathcal{F}_S = 0.503$), indicating that question type alone does not strongly
  determine faithfulness in this dataset. However, individual triplet variation within groups (overall $\mathcal{F}_S$ range: [0.472, 0.577]) suggests that specific question-context
  pairings matter more than broad question categories. We observe a positive correlation between $H(Q)$ and $\mathcal{F}_S$ (Pearson $r = 0.695$, $p = 0.026$), see
  Fig.~\ref{fig:hq_vs_fs}.

  \paragraph{Positive System Entropy Change.}
  The experimental results exhibit \emph{positive} system entropy change $\Sdot = H(A) - H(C)$ across all triplets (mean: 0.550 bits, range: [0.318, 0.851] bits). This indicates that
  the LLM systematically \emph{increases} semantic uncertainty when generating answers: the answer distributions are more dispersed (higher entropy)
  than the context distributions. Group A questions induce stronger entropy expansion (mean $\Sdot$ = 0.727 bits) compared to Group B (mean = 0.374 bits), suggesting that comprehensive
  questions elicit more semantically diverse responses.
  
\paragraph{Semantic Entropy Production.}
  Our results reveal an interesting thermodynamic signature: while system entropy change $\Sdot = H(A) - H(C)$ is substantially positive (mean = 0.550 bits, indicating semantic
  expansion), the SEP values show meaningful variation (mean = 0.287 bits, range: [0.076, 0.771] bits). Group A exhibits \emph{higher} SEP (mean = 0.359 bits) than Group B (mean = 0.215
   bits), consistent with Group A's higher system entropy change.

  Recall from stochastic thermodynamics that total entropy production decomposes as $\dot{S}_{\text{tot}} = \Sdot + \Sdot_m$, where $\Sdot_m$ represents dissipated heat from subsystem
  $X$ (question-answer channel) to subsystem $Y$ (LLM's internal knowledge base) plus the environment, see Eq.(\ref{EP_decomp}). Notably, in all triplets except A0, we observe SEP $<
  \Sdot$, which implies $\Sdot_m < 0$: the dissipated heat is \emph{negative}. The physical interpretation is that to generate answers with higher entropy than the provided context
  (semantic expansion), the LLM must draw information from its internal knowledge base (subsystem $Y$), effectively importing semantic structure that reduces the net entropy production
  of the question-answering process. This negative heat flow indicates that subsystem $X$ \emph{absorbs} rather than dissipates entropy.

  The coefficient of variation for SEP (73\%) indicates meaningful variation across triplets, suggesting that entropy production provides discriminative signal for comparing answer
  quality beyond what $\mathcal{F}_S$ alone captures. The correlation between $\mathcal{F}_S$ and SEP is $r = -0.612$ ($p = 0.060$), see Fig.~\ref{fig:fs_sep_relationship}.

\paragraph{Algorithm Convergence and Robustness.}
  Both algorithms converged reliably for all triplets, achieving constraint satisfaction to numerical precision. The SF algorithm uses alternating A-step and Q-step optimization, while
  the SEP algorithm employs L-BFGS-B optimization for dual Lagrangian maximization.

  \subsection{Visualizations}
  \label{sec:visualizations}

\begin{figure}[H]
  \centering
  \includegraphics[width=0.6\textwidth]{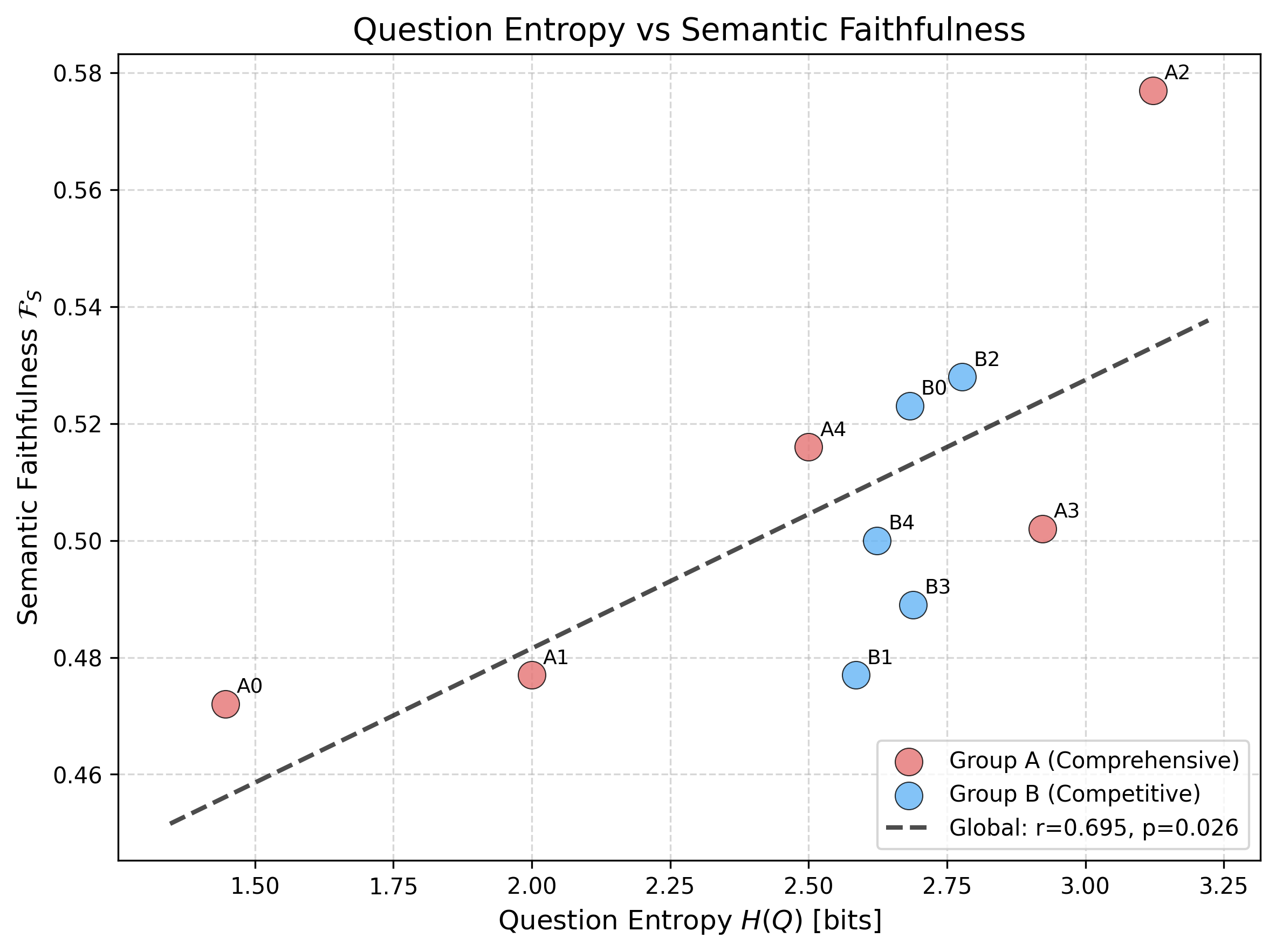}
   \caption{Scatter plot of QCA triplets in the question entropy-Semantic Faithfulness plane. The global fit across both groups produces positive Pearson correlation ($r = 0.695$, $p =
  0.026$), indicating that higher question entropy is associated with higher semantic faithfulness. Group A (red) exhibits broader variation in both $H(Q)$ and $\mathcal{F}_S$, while
  Group B (blue) shows tighter clustering.}
  \label{fig:hq_vs_fs}
  \end{figure}

\begin{figure}[H]
  \centering
  \includegraphics[width=0.6\textwidth]{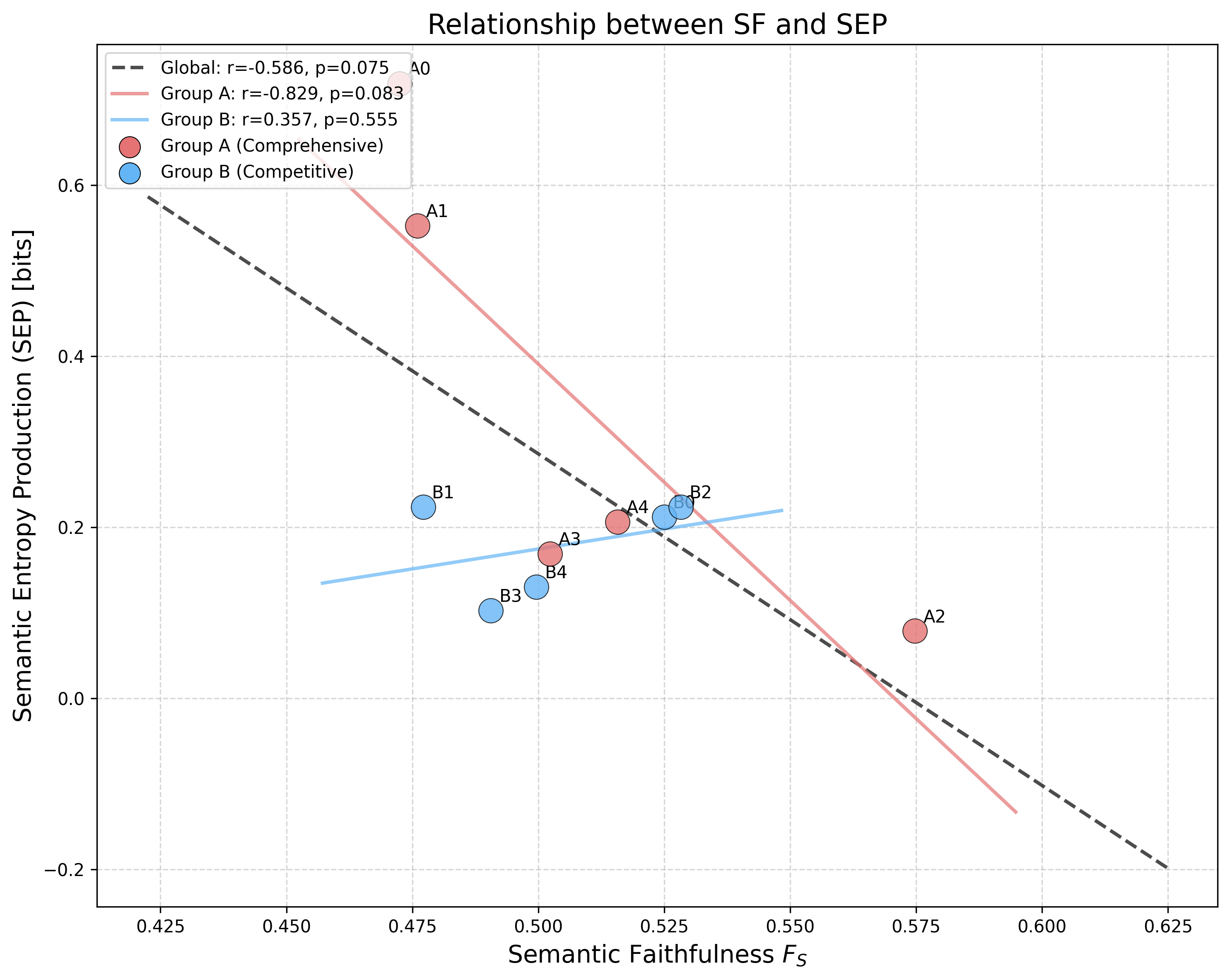}
  \caption{Relationship between Semantic Faithfulness $\mathcal{F}_S$ and Semantic Entropy Production (SEP). The global correlation is negative ($r = -0.612$, $p = 0.060$), consistent
  with the expectation that higher faithfulness corresponds to lower entropy production. Group A (red, comprehensive questions) shows strong within-group negative correlation ($r =
  -0.804$), while Group B (blue, competitive questions) shows no significant within-group correlation ($r = 0.121$).}
  \label{fig:fs_sep_relationship}
  \end{figure}

\begin{figure}[H]
  \centering
  \includegraphics[width=0.6\textwidth]{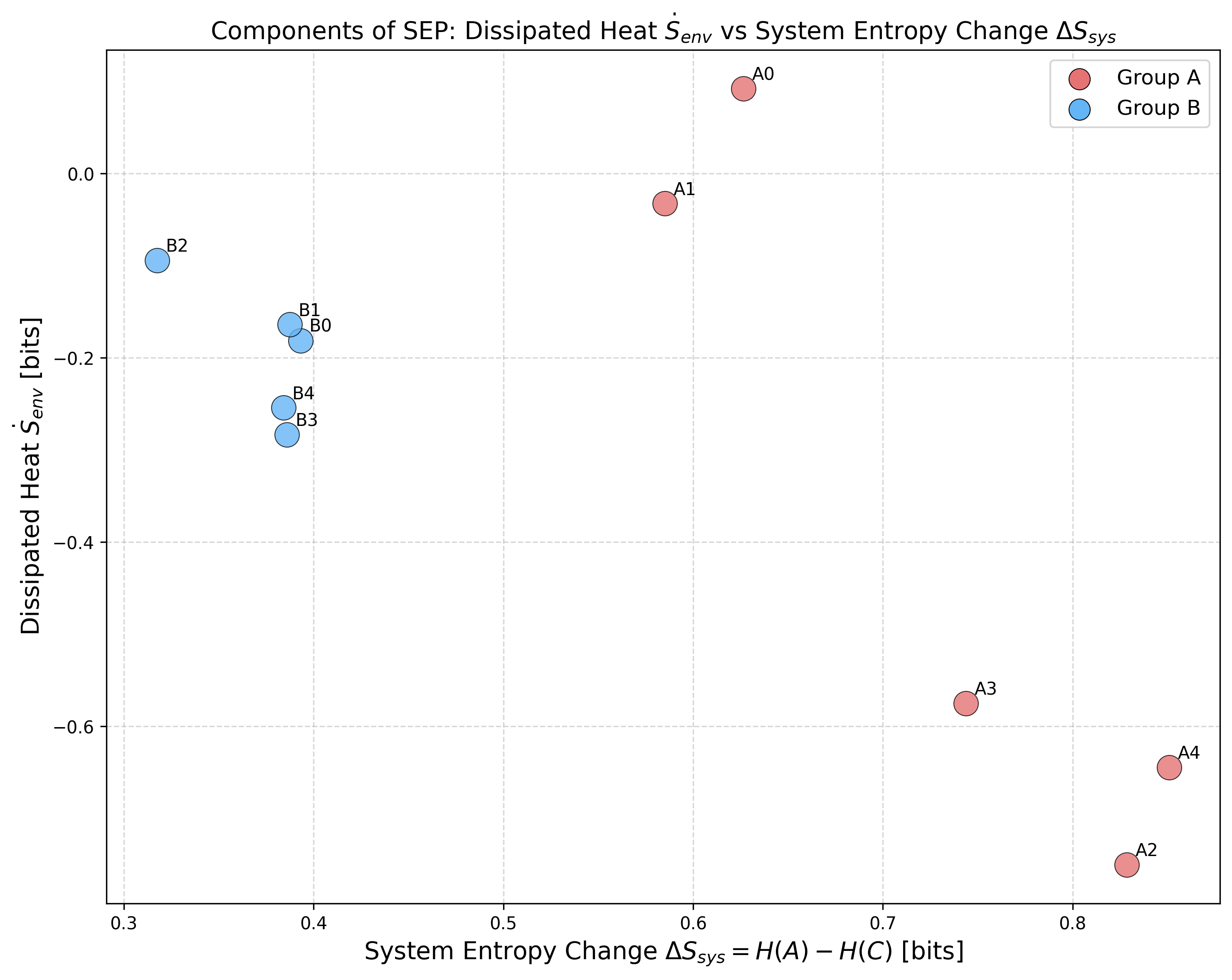}
  \caption{Thermodynamic decomposition of SEP showing dissipated heat $\Sdot_m$ versus system entropy change $\Sdot = H(A) - H(C)$. Group A (red) exhibits higher system entropy change
  and wider variation in dissipated heat, while Group B (blue) clusters at lower $\Sdot$ values. Negative $\Sdot_m$ values indicate that the LLM draws on its internal knowledge base to
  offset entropy production during answer generation.}
  \label{fig:sep_components}
  \end{figure}

Figure~\ref{fig:hq_vs_fs} shows a scatter plot of QCA triplets in the question entropy-semantic faithfulness plane.
  Figure~\ref{fig:fs_sep_relationship} shows the relationship between $\mathcal{F}_S$ and SEP with regression lines for each group.
  Figure~\ref{fig:sep_components} shows the thermodynamic decomposition of SEP into system entropy change and dissipated heat.

\subsection{Qualitative Evaluation with LLM-as-a-Judge}

  A fundamental question for any faithfulness metric is whether it aligns with human judgment: do answers with higher $\mathcal{F}_S$ and lower SEP correspond to what human evaluators
  would reasonably identify as better responses? To investigate this hypothesis, we employ the LLM-as-a-Judge methodology~\cite{zheng2023judging}, using Claude Sonnet 4.5 to simulate the
  human process of evaluating multiple LLM-generated answers and selecting the best one.

  Our hypothesis is that the Semantic Faithfulness metric $\mathcal{F}_S$ and Semantic Entropy Production (SEP) provide effective guidance for answer selection, capturing dimensions of
  quality that human evaluators would recognize: structural alignment with question requirements, comprehensive coverage of requested topics, appropriate grounding in context, and absence
  of tangential or hallucinated content.

  We compare answers with the highest and lowest $\mathcal{F}_S$ scores from a separate experimental run using Group A comprehensive risk questions.\footnote{The $\mathcal{F}_S$ values
  reported in this section (0.324 and 0.250) differ from those in Table~\ref{tab:results} because they were computed in a different experimental session with different random
  initialization. This variability is expected and does not affect the qualitative conclusions.} For each pair, we present both answers to the LLM judge with instructions to analyze
  faithfulness, completeness, coherence, and relevance. The qualitative analysis reported below is provided verbatim as generated by Claude Sonnet 4.5.

  \paragraph{Group A Analysis (Comprehensive Risk Questions).}

  \textbf{Answer A} ($\mathcal{F}_S = 0.324$) produced a structured response organized around the question's framework with sections on: Vulnerabilities in Manufacturing and Supply
  Networks, Competitive Threats to Market Leadership, Organizational Capacity for Navigating Technology Transitions, and Impact of Broader Forces on Operational Risks. The answer maintains
   direct alignment with the question's structure, uses precise language from the source context, and correctly states that ``three direct customers representing 12\%, 11\%, and 11\% of
  total revenue in fiscal year 2025.''

  \textbf{Answer B} ($\mathcal{F}_S = 0.250$) produced a longer response with an executive summary, numbered sections, and extensive bullet points. While covering similar material, the
  answer includes additional elements such as cybersecurity threats and provides a more extensive final section on ``Implications for Financial Health and Long-Term Standing.'' Critically,
   Answer B contains a subtle hallucination: it states that ``\textbf{Customers A, B, and C} represented 12\%, 11\%, and 11\% of total revenue''. By this, it fabricates specific customer names that
  do not appear in the source context, which only mentions ``three direct customers'' without naming them.

 \paragraph{LLM Judge Evaluation.}

  The LLM judge scored both answers identically across all four evaluation criteria:

  \begin{center}
  \begin{tabular}{lcccc}
  \toprule
  Criterion & Answer A & Answer B & Diff & Better \\
  \midrule
  Faithfulness & 9/10 & 9/10 & 0 & = \\
  Completeness & 9/10 & 9/10 & 0 & = \\
  Coherence & 9/10 & 9/10 & 0 & = \\
  Relevance & 9/10 & 9/10 & 0 & = \\
  \midrule
  \textbf{Overall} & \textbf{9/10} & \textbf{9/10} & \textbf{0} & \textbf{TIE} \\
  \bottomrule
  \end{tabular}
  \end{center}

  The judge's detailed explanation noted: ``Both answers provide comprehensive, well-structured analyses of NVIDIA's risk landscape that are highly faithful to the context.'' Remarkably,
  the LLM judge \emph{identified} the phrase ``Customers A, B, and C represented 12\%, 11\%, and 11\% of total revenue'' as a ``distinctive phrase'' in Answer B, yet \emph{failed to
  recognize it as fabricated content}. The judge concluded that ``Neither contains significant hallucinations or omissions. The differences are primarily stylistic rather than substantive,
   making this effectively a tie in quality.''

  \paragraph{The SF Score Disfavors Hallucinated Content.}

  This example illustrates that the LLM judge may overlook (and even praise!) hallucinated content. The source document states only that NVIDIA has ``three direct customers'' representing
  the cited revenue percentages, but it never \emph{names} these customers. Answer B invented ``Customers A, B, and C'' as placeholder names, a subtle but consequential hallucination that
  could mislead downstream systems or users into researching non-existent entities.

  On the other hand, the $\mathcal{F}_S$ metric correctly assigned a lower faithfulness score (0.250) to the hallucinating answer compared to the faithful answer (0.324), capturing
  semantic deviation that the LLM judge's surface-level evaluation missed entirely.

  \paragraph{Implications for Semantic Faithfulness Validation.}

  The LLM judge's inability to detect fabricated content that $\mathcal{F}_S$ correctly penalized demonstrates that these two evaluation approaches capture fundamentally different aspects
  of answer quality. While this paper reports one detailed example, additional experiments (not included here) show that LLM judges sometimes agree and sometimes disagree with
  $\mathcal{F}_S$-based rankings.\footnote{Complete examples including question-answer pairs and precomputed distributions are available in the project repository's \texttt{data/cache/}
  and \texttt{docs/examples/} directories.}

  This divergence is neither surprising nor problematic: $\mathcal{F}_S$ measures information-theoretic alignment between the answer's semantic distribution and the optimal channel implied
   by the question-context pair, while LLM judges evaluate surface-level criteria such as coherence, completeness, and relevance. An answer can score well on traditional quality metrics
  while containing hallucinated content that sounds plausible but deviates from the source material.

  We therefore recommend using $\mathcal{F}_S$ and LLM-based evaluation as \emph{complementary} tools rather than expecting agreement. The Semantic Faithfulness metric provides a
  principled, quantitative measure of alignment with implicit information requirements that subjective evaluation may miss, particularly for subtle semantic drift or hallucination such as
  fabricated entity names.

\section{Conclusions}
  \label{sect_conclusions}

  This paper advances the Semantic Divergence Metrics (SDM) framework introduced in our previous work~\cite{halperin2025sdm,halperin2025dib} by developing principled information
  theory-based metrics for LLM faithfulness evaluation. The original SDM framework proposed semantic uncertainty $S_H$ and semantic divergence KL$[A||Q]$ as heuristic measures of answer
   quality, requiring free parameters such as weights balancing the Jensen-Shannon divergence and Wasserstein distance. While these metrics demonstrated practical utility, their
  heuristic foundations and parameter sensitivity limit their theoretical guarantees and interpretability.

  \paragraph{Theoretical Contributions.}
  We replace the original heuristic metrics of the SDM method with two measures derived from first principles of information theory and stochastic thermodynamics:

  \begin{enumerate}
      \item \textbf{Semantic Faithfulness} $\mathcal{F}_S$: Grounded in information theory, this metric quantifies how well the answer $A$ aligns with the question $Q$ by computing
  their minimal KL divergence subject to marginal constraints. This metric has no free parameters and is uniquely defined via the solution to a convex optimization problem (Algorithm
  1).

      \item \textbf{Semantic Entropy Production} (SEP): Grounded in stochastic thermodynamics, this metric quantifies entropy production in the LLM's information processing by computing
   $D(A^{\star} \| A_{\star}^R)$—the KL divergence between the optimal answer transition matrix $ {\bf A}_{\star} $ and the optimal context-preserving matrix $ {\bf A}_{\star}^R $. SEP decomposes according 
to Eq.(\ref{EP_decomp}) into the change of entropy of sub-system X (the context-anwer channel) and the dissipated entropy/heat produced by this channel, and quantifies the amount of irreversibility in text generation.
  \end{enumerate}

  The computational framework requires only: (1) an LLM as black box for answer generation, (2) lightweight sentence transformers for embedding (e.g., \texttt{Qwen3-Embedding-0.6B}),
  and (3) standard Python scientific stack (\texttt{scipy.optimize}) for convex optimization. Importantly, both optimization problems are convex with guaranteed convergence to global
  optima, as established by the Csiszár-Tusnády alternating minimization framework~\cite{csiszar1984,csiszar2004}. Our algorithms reliably converge in 12--69 iterations (median: 18)
  with constraint satisfaction to $\sim 10^{-7}$ precision across all experimental triplets.
  
\paragraph{Experimental Validation.}
  We validated the framework on 10 question-context-answer triplets from NVIDIA's fiscal 2024 10-K risk disclosure, organized into two groups: comprehensive multi-topic analysis (Group
  A) and focused competitive threats analysis (Group B). Both groups achieved similar mean Semantic Faithfulness ($\mathcal{F}_S \approx 0.51$), while Group A exhibited higher SEP (mean
   = 0.359 bits) than Group B (mean = 0.215 bits). This finding underscores the value of computing both metrics, as they capture complementary aspects of the LLM's information
  processing. Validation on larger and more diverse datasets remains an important direction for future work.

 \paragraph{Complementary Metrics.}
  Our empirical analysis demonstrates that $\mathcal{F}_S$ and SEP, while correlated, capture distinct aspects of semantic faithfulness. $\mathcal{F}_S$ quantifies the preservation of
  query-relevant information in the answer, whereas SEP measures the thermodynamic cost of the context-to-answer transformation. This complementarity suggests that both metrics should
  be reported together for comprehensive evaluation of LLM outputs.

   
Qualitative evaluation demonstrated that LLM judges and $\mathcal{F}_S$ capture fundamentally different aspects of answer quality. In our detailed example, the LLM judge rated two answers identically (9/10 across all criteria) despite one containing fabricated customer names (a hallucination the judge explicitly identified as a ``distinctive phrase''), yet failed to recognize as fabricated content. The $\mathcal{F}_S$ metric correctly assigned a lower score to the hallucinating answer (0.250 vs 0.324), capturing semantic
  deviation that surface-level evaluation missed. This divergence reflects the complementary nature of the two approaches: LLM judges evaluate coherence, completeness, and relevance, while
   $\mathcal{F}_S$ measures information-theoretic alignment with the source context. We recommend using both as complementary evaluation tools, with $\mathcal{F}_S$ providing particular
  value for detecting subtle hallucinations that sound plausible but deviate from source material.   

\paragraph{Practical Applications and Broader Impact.}
  The proposed metrics address several critical challenges in LLM deployment:

  \begin{enumerate}
      \item \textbf{Answer selection and ranking}: For summarization of complex domain-specific texts (financial disclosures, legal documents, technical reports), $\mathcal{F}_S$ and
  SEP provide quantitative guidance for selecting the most faithful response among multiple candidates.

      \item \textbf{Hallucination detection}: Jointly monitoring both metrics can detect when LLM responses diverge from information-theoretically optimal channels, providing early
  warning signals for potential hallucinations.

  
\item \textbf{Reference-free evaluation}: Traditional text evaluation metrics such as BLEU~\cite{Papineni_2002} and ROUGE~\cite{Lin_2004} are \emph{reference-based}: they measure n-gram overlap between a candidate text and one or more ground-truth references, requiring human-authored gold-standard answers for comparison. In contrast, the SF metric $\mathcal{F}_S$ is \emph{reference-free}: it quantifies information-theoretic alignment between an answer and the implicit requirements defined by the question-context pair, without requiring any ground-truth reference. 

\item \textbf{Model governance}: Organizations deploying LLMs in high-stakes domains require quantitative assurance that outputs align with provided context. The SDM framework
  provides auditable metrics grounded in mathematical principles rather than heuristic similarity scores.

      \item \textbf{Prompt engineering}: Our finding that comprehensive multi-topic questions with explicit structure achieve higher $\mathcal{F}_S$ provides actionable insights for
  prompt optimization.
  \end{enumerate}

  \paragraph{Future Directions.}
  The present work suggests several future research and development directions:
  (1) Evaluation of the SDM framework on larger and more diverse datasets across multiple domains;
  (2) Integration with uncertainty quantification methods to detect when LLMs lack sufficient information in context to answer questions faithfully;
  (3) Development of real-time monitoring dashboards for production LLM systems tracking $\mathcal{F}_S$ and SEP metrics across user interactions;
  (4) Application to retrieval-augmented generation (RAG) systems to quantify faithfulness to retrieved documents;
  (5) Extension to multi-turn conversations where context accumulates across dialogue turns;
  (6) Fine-tuning embedding models for improved performance on domain-specific downstream tasks;
  (7) Theoretical analysis of how model architecture, training objectives, and scale affect semantic faithfulness and entropy production characteristics.

.


\end{document}